\documentclass[acmsmall,screen,nonacm]{acmart}

\usepackage{multirow}

\AtBeginDocument{%
  \providecommand\BibTeX{{%
    \normalfont B\kern-0.5em{\scshape i\kern-0.25em b}\kern-0.8em\TeX}}}

\setcopyright{acmcopyright}
\copyrightyear{2018}
\acmYear{2018}
\acmDOI{10.1145/1122445.1122456}

\acmConference[Woodstock '18]{Woodstock '18: ACM Symposium on Neural
  Gaze Detection}{June 03--05, 2018}{Woodstock, NY}
\acmBooktitle{Woodstock '18: ACM Symposium on Neural Gaze Detection,
  June 03--05, 2018, Woodstock, NY}
\acmPrice{15.00}
\acmISBN{978-1-4503-XXXX-X/18/06}

\usepackage{xparse}
\usepackage{booktabs}
\newlength{\NFwidth}
\NewDocumentCommand{\NFelement}{mmm}{\normalsize\textbf{#1} #2\hfill #3}
\NewDocumentCommand{\NFline}{O{l}m}{\footnotesize\makebox[\NFwidth][#1]{#2}}

\NewDocumentCommand{\NFentry}{sm}{%
  \makebox[.5\NFwidth][l]{\normalsize
    \IfBooleanT{#1}{\makebox[0pt][r]{\textbullet\ }}%
    #2}\ignorespaces}
\NewDocumentCommand{\NFtext}{+m}
 {\parbox{\NFwidth}{\raggedright#1}}

\newcommand{\NFtitle}{\multicolumn{1}{c}{\huge\bfseries Dataset Facts}}

\newcommand{\NFRULE}{\midrule[5pt]}
\newcommand{\NFRule}{\midrule[2pt]}
\newcommand{\NFrule}{\midrule}

\begin{document}

\title{Addressing "Documentation Debt" in Machine Learning Research: A Retrospective Datasheet for BookCorpus}


\author{Jack Bandy}
\author{Nicholas Vincent}


\begin{abstract}
Recent literature has underscored the importance of dataset documentation work for machine learning, and part of this work involves addressing ``documentation debt'' for datasets that have been used widely but documented sparsely. This paper aims to help address documentation debt for BookCorpus, a popular text dataset for training large language models. Notably, researchers have used BookCorpus to train OpenAI's GPT-N models and Google's BERT models, even though little to no documentation exists about the dataset's motivation, composition, collection process, etc. We offer a preliminary datasheet that provides key context and information about BookCorpus, highlighting several notable deficiencies. In particular, we find evidence that (1) BookCorpus likely violates copyright restrictions for many books, (2) BookCorpus contains thousands of duplicated books, and (3) BookCorpus exhibits significant skews in genre representation. We also find hints of other potential deficiencies that call for future research, including problematic content, potential skews in religious representation, and lopsided author contributions. While more work remains, this initial effort to provide a datasheet for BookCorpus adds to growing literature that urges more careful and systematic documentation for machine learning datasets.
\end{abstract}




\maketitle

\section{Introduction}
Large language models are ``growing'' in a number of ways: the number of parameters in the models (e.g. 175 Billion in OpenAI's full GPT-3 \cite{brown2020language}), the range of use cases (e.g. to help train volunteer counselors for the Trevor Project \cite{ohleiser2021trevorproject}), the degree to which these models affect the public (e.g. powering almost every English query on Google \cite{schwartz2020bert}), and crucially, the size and complexity of the text data used for training. 
Bender and Gebru et al. 
\cite{bender2021dangers} suggest that training data currently faces ``documentation debt,'' in that popular language models are trained on sparsely-documented datasets which are often difficult to replicate and comprehend.

One such sparsely-documented dataset is BookCorpus. Originally introduced by Zhu and Kiros et al. \cite{zhu2015aligning} in 2014, BookCorpus and derivative datasets have been used to train Google's massively influential ``BERT'' model \cite{devlin2018bert} (amassing over 17,000 Google Scholar citations as of April 2021), BERT's variants such as RoBERTa \cite{liu2019roberta} and ALBERT \cite{lan2019albert}, OpenAI's GPT-N models \cite{radford2019language}, XLNet \cite{yang2019xlnet}, and more. Yet researchers provide scant details about BookCorpus, often merely noting the number of books and tokens in the dataset, or the total disk space it occupies. When introducing the dataset in 2014, \citet{zhu2015aligning} provided six summary statistics (shown in Table \ref{table-zhu}) along with the following description:

\begin{quote}
In order to train our sentence similarity model we collected a corpus of 11,038 books from the web. These are free books written by yet unpublished authors. We only included books that had more than 20K words in order to filter out perhaps noisier shorter stories. The dataset has books in 16 different genres, e.g., Romance (2,865 books), Fantasy (1,479), Science fiction (786), etc. Table [1] highlights the summary statistics of our corpus.
\end{quote}

\begin{table}[h]
\resizebox{1.0\textwidth}{!}{
\begin{tabular}{|c|c|c|c|c|c|}
\hline
\# of books & \# of sentences & \# of words & \# of unique words & mean \# of words per sentence & median \# of words per sentences \\ \hline
11,038      & 74,004,228      & 984,846,357 & 1,316,420          & 13                            & 11                               \\ \hline
\end{tabular}
}
\caption{The six summary statistics of BookCorpus originally provided in Table 2 from \citet{zhu2015aligning}}
\label{table-zhu}
\end{table}

Our paper attempts to help address ``documentation debt'' for the widely-used BookCorpus dataset, following a growing body of work suggesting that influential datasets need more careful documentation (i.e. ``dataset nutrition labels'' \cite{holland2020dataset,holland2018dataset,sun2019mithralabel}, ``data statements'' \cite{bender2018data}, ``dataset disclosure forms'' \cite{shope2021lawyer}, or ``datasheets'' \cite{gebru2018datasheets,bender2021dangers}). Ongoing findings underscore the importance of data documentation. For example, Northcutt et al. \cite{northcutt2021pervasive} found pervasive label errors in test sets from popular benchmarks used in computer vision (e.g. MNIST, ImageNet) and natural language processing (e.g. IMDB movie reviews, 20 Newsgroups, Amazon Reviews). Building on this work, we provide documentation for an unlabeled dataset that has mainly been used in unsupervised settings (such as training large language models). We apply the datasheet framework described by \citet{gebru2018datasheets} to retrospectively document the motivation, composition, and collection process for BookCorpus, as well as other aspects for researchers to consider before using the dataset.

A major challenge in this effort is that \textit{there is no official public version of BookCorpus}. There have been several efforts, such as BookCorpusOpen \cite{bookcorpusopencard}, which replicate BookCorpus using the same approach used in the original paper (scraping free books from smashwords.com). In our efforts to create a datasheet, we consider three versions: the original 2014 BookCorpus (collected from the authors' website \cite{zhu2015aligning}), BookCorpusOpen \cite{bookcorpusopencard} (a 2020 version included in a dataset collection called ``The Pile'' \cite{gao2020pile}), and Smashwords21 (a ``superset'' of all books listed on smashwords.com, which we collected ourselves). Note that Smashwords21 only includes metadata about each book in the set, and not the full text of each book (non-free books would cost over \$1 million USD to purchase and download based on the metadata we collected).

In addition to documenting several important aspects of BookCorpus such as its original use cases and sources of funding, we find several notable deficiencies in the dataset. For one, we find that many books contain copyright restrictions that should have prevented them from being distributed in BookCorpus and similar datasets. We also find that thousands of books in BookCorpus are duplicated, with only 7,185 unique books out of 11,038 total. Third, we find notable genre skews in the dataset, for example romance novels are significantly over-represented compared to the newer BookCorpusOpen as well as the Smashwords21 superset. In addition to these three deficiencies, we also find a number of \textit{potential} deficiencies that motivate future work: Smashwords21 points to a range of potentially problematic content, skewed religious representation, and lopsided author contributions (with some super-authors publishing thousands of books). We conclude with a discussion of future work, implications, and the value of documentation for machine learning research.


\section{Methods}

\subsection{Documentation and Analysis}
The authors systematically addressed all questions suggested by \cite{gebru2018datasheets} for creating datasheets.\footnote{Replication materials (data and code) are available at  \url{https://github.com/jackbandy/bookcorpus-datasheet}} While we did not deem all questions relevant to BookCorpus, for transparency, we still include these questions and note our reasoning. Furthermore, as encouraged by \cite{gebru2018datasheets}, we include some additional questions that are important for understanding and using BookCorpus. To distinguish between ``official'' datasheet questions from \cite{gebru2018datasheets} and additional questions that we added, our additions are denoted with a \textit{[+]} preceding the question.

\subsection{Data Collection}
To create this datasheet, we collected and analyzed three different versions of the dataset: (1) the original 2014 BookCorpus (collected from the authors' website \cite{zhu2015aligning}), (2) BookCorpusOpen \cite{bookcorpusopencard} (a 2020 version included in a dataset called ``The Pile'' \cite{gao2020pile}), and (3) Smashwords21 (a ``superset'' of all books listed on smashwords.com, which we collected ourselves).

\subsubsection{Original BookCorpus Dataset}\label{sec-original-bc}
While BookCorpus is no longer publicly available, we obtained copies of the dataset files directly from the authors' website\footnote{We have notified the authors of the security vulnerability that allowed us to download the dataset.} where it was previously distributed \cite{zhu2015aligning}. Specifically, we obtained a directory called \texttt{books\_txt\_full} that contains 16 folders corresponding to the genres of the books (e.g. Adventure, Fantasy, Historical, etc.), as well as a directory called \texttt{books\_in\_sentences} that contains two large files (\texttt{books\_large\_p1.txt} and \texttt{books\_large\_p2.txt}) with one sentence per line.

Within the \texttt{books\_in\_sentences} directory, the \texttt{books\_large\_p1.txt} file contained 536,517,284 words, and \texttt{books\_large\_p1.txt} contained 448,329,073 (based on the ``-wc'' unix word count software), for a total of 984,846,357 words. This aligns exactly with the number of words reported by \cite{zhu2015aligning} when introducing BookCorpus. Also, the first file contained 40,000,000 sentences, and the second file contained 34,004,228, together accounting for the 74,004,228 sentences reported by \cite{zhu2015aligning} when introducing BookCorpus.

The \texttt{books\_txt\_full} directory contained 11,040 text files, even though \cite{zhu2015aligning} report 11,038 books in the original BookCorpus dataset. Two extra files account for the discrepancy: one called \texttt{romance-all.txt} (1.12 GB) and another called \texttt{adventure-all.txt} (150.9 MB), large files which appear to contain concatenated text from all books within the respective genres.

However, the individual text files from the \texttt{books\_txt\_full} directory only contained 811,601,031 total words (after removing the \texttt{romance-all.txt} and \texttt{adventure-all.txt} files) -- more than 170M words shy of the full sentence corpus. This is likely due in part to some empty files in the data we obtained (e.g. \texttt{Magic\_of\_the\_Moonlight.txt} and \texttt{Song-of-Susannah.txt}), although we have no way to identify the complete causes of the word count discrepancy.

\subsubsection{BookCorpusOpen Dataset}
For the purposes of comparison, we also downloaded a newer, replicated version of BookCorpus which we refer to as \textit{BookCorpusOpen}, in line with a publicly-maintained version of the dataset \cite{bookcorpusopencard}. BookCorpusOpen is included in the Pile dataset as BookCorpus2 \cite{gao2020pile} and has been referred to by various other names (e.g. BookCorpusNew, Books1, OpenBookCorpus). The files we inspect include a list of 18,060 URLs from smashwords.com, and corresponding text files for 17,868 of them.

\subsubsection{Smashwords21 ``Superset''}
To help address questions related to sampling, we collected a superset of the books represented in BookCorpus and BookCorpusOpen. Originally, BookCorpus contained all free English books from smashwords.com which were longer than 20,000 words. A ``complete'' superset might contain all books ever published, or a similarly vast collection. For the purposes of this paper, we collected metadata about 411,826 unique books published on smashwords.com as of April 2021.

To create this superset, we scraped all books listed on smashwords.com, similar to what has been done in efforts to replicate BookCorpus \cite{soskkobayashi2018bookcorpus}. Notably, our scrape recovered 411,826 unique books, while Smashwords reported that over 565,000 total books had been published on the website at the time. This discrepancy likely stems from a default filter that excludes adult erotica from the public listings. We could have set a \texttt{no\_filtering} cookie in our scraping program to include these books, however, BookCorpus and BookCorpusOpen do not mention this, so we only scraped books that were publicly-listed by default. We ran the scraping program twice to help ensure coverage.


\section{Summary Data Card for BookCorpus}
\vspace{0.07in}
\sffamily
{\centering
\fbox{%
\begin{tabular}{@{}p{\NFwidth}@{}}
\NFtitle\\
\NFrule
\NFtext{\textbf{Dataset} BookCorpus}\\

\NFtext{\textbf{Instances Per Dataset} 7,185 unique books, 11,038 total}\\
\NFRULE
\NFline{Motivation}\\
\NFrule

\NFelement{Original Authors}{}{Zhu and Kiros et al. (2015) \cite{zhu2015aligning}}\\
\NFelement{Original Use Case}{}{Sentence embedding }\\
\NFelement{Funding}{}{Google, Samsung, NSERC, CIFAR, ONR}\\

\NFRule
\NFline{Composition}\\
\NFrule

\NFelement{Sample or Complete}{}{Sample, $\approx$2\% of smashwords.com in 2014}\\

\NFelement{Missing Data}{}{98 empty files, $\leq$655 truncated files}\\

\NFelement{Sensitive Information }{}{Author email addresses}\\

\NFRule
\NFline{Collection}\\
\NFrule
\NFelement{Sampling Strategy}{}{Free books with $\geq$20,000 words}\\
\NFelement{Ethical Review}{}{None stated}\\
\NFelement{Author Consent}{}{None}\\

\NFRule
\NFline{Cleaning and Labeling}\\
\NFrule
\NFelement{Cleaning Done}{}{None stated, some implicit}\\
\NFelement{Labeling Done}{}{None stated, genres by smashwords.com}\\

\NFRule
\NFline{Uses and Distribution}\\
\NFrule
\NFelement{Notable Uses}{}{Language models (e.g. GPT \cite{radford2019language}, BERT \cite{devlin2018bert}})\\
\NFelement{Other Uses}{}{List available on HuggingFace \cite{bookcorpuscard}}\\
\NFelement{Original Distribution}{}{Author website (now defunct) \cite{zhu2015aligning}}\\
\NFelement{Replicate Distribution}{}{BookCorpusOpen \cite{bookcorpusopencard}}\\

\NFRule
\NFline{Maintenance and Evolution}\\
\NFrule
\NFelement{Corrections or Erratum}{}{None}\\
\NFelement{Methods to Extend}{}{``Homemade BookCorpus'' \cite{soskkobayashi2018bookcorpus}}\\
\NFelement{Replicate Maintainers}{}{Shawn Presser \cite{bookcorpuscard}}\\

\NFRULE
\NFline{Genres \hfill\% of BookCorpus*}\\
\NFrule
\NFelement{Romance}{2,881\,books}{26.1\%}\\
\NFrule
\NFelement{Fantasy}{1,502\,books}{13.6\%}\\
\NFrule
\NFelement{Vampires}{600\,books}{5.4\%}\\
\NFrule
\NFentry{Horror 4.1\%}
\NFentry*{Teen 3.9\%}\\
\NFentry{Adventure 3.5\%}
\NFentry*{Literature 3.0\%}\\
\NFentry{Historical Fiction 1.6\%}\\
\NFrule
\NFtext{Not a significant source of nonfiction.}\\
\NFrule
\NFtext{* Percentages based on directories in books\_txt\_full. Some books cross-listed.}
\end{tabular}}

\par} 

\section{Full Datasheet for BookCorpus}

\subsection{Motivation}
\subsubsection{For what purpose was BookCorpus created?}
BookCorpus was originally created to help train a neural network that could provide ``descriptive explanations for visual content'' \cite{zhu2015aligning}. Specifically, BookCorpus trained a sentence embedding model for aligning dialogue sentences from movie subtitles with written sentences from a corresponding book. After unsupervised training on BookCorpus, the authors' encoder model could ``map any sentence through the encoder to obtain vector representations, then score their similarity through an inner product'' \cite{zhu2015aligning}.

\subsubsection{[+] For what purpose were the books in BookCorpus created?}
The books in BookCorpus were self-published by authors on smashwords.com, likely with a range of motivations. While we can safely assume that authors publishing free books via smashwords.com had some motivation to share creative works with the world, there is no way to verify they were interested in training AI systems. For example, many authors in BookCorpus explicitly license their books ``for [the reader's] personal enjoyment only,'' limiting reproduction and redistribution. When notified about BookCorpus and its uses, one author from Smashwords said ``it didn’t even occur to me that a machine could read my book'' \cite{lea2016google}.

\subsubsection{Who collected BookCorpus?}
BookCorpus was collected by Zhu and Kiros et al. \cite{zhu2015aligning} from the University of Toronto and the Massachusetts Institute of Technology. Their original paper includes seven authors, but does not specify who was involved in collecting BookCorpus.

\subsubsection{[+] Who created the books in BookCorpus?}
BookCorpus' constituent data was created by a large number of self-published authors on smashwords.com. These authors wrote the books and sentences that make up BookCorpus, and now support a wide range of machine learning systems.

\subsubsection{[+] How many people were involved in creating BookCorpus?}

It is challenging to estimate the exact number of authors who contributed to the original BookCorpus, as the dataset does not provide structured metadata. Instead, we provide an estimate based on the number of unique authors who contributed free books to Smashwords21. In Smashwords21, 29,272 unique authors contributed 65,556 free books, which included 1.77 billion total words. Assuming a similar ratio (unique authors:free books) in the original BooksCorpus, we estimate that about 3,490 authors were involved in creating  the original dataset of 7,185 books (29,272 / 65,556 * 7,815 = 3,489.5). Author contributions also appear to be highly concentrated: among free books in Smashwords21, the top 10\% of authors by word count were responsible for 59\% of all words in the dataset, and the top 10\% by book count were responsible for 43\% of all books.

\subsubsection{Who funded the creation of BookCorpus?}
The original paper by Zhu and Kiros et al. \cite{zhu2015aligning} acknowledges support from the Natural Sciences and Engineering Research Council (NSERC), the Canadian Institute for Advanced Research (CIFAR), Samsung, Google, and a grant from the Office of Naval Research (ONR). They do not specify how funding was distributed across these sources.


It is more difficult to identify funding for the authors who wrote the books in BookCorpus. Broadly, many authors on Smashwords do make money by selling ebooks to readers (including on other platforms like Kindle, Audible, Barnes and Noble, and Kobo), although many also write books as a hobby alongside other occupations. Some books in BookCorpus may have been commissioned in some way, however, analyzing sources of commission would require further work.

\subsection{Composition}
\subsubsection{What do the instances in BookCorpus represent?}
BookCorpus consists of text files, each of which corresponds to a single book from smashwords.com. Zhu and Kiros et al. \cite{zhu2015aligning} also provide two large files in which each row represents a sentence. 

\subsubsection{How many instances (books) are there in total?}
In the original dataset described by Zhu and Kiros et al. \cite{zhu2015aligning}, BookCorpus contained \textbf{11,038} books. However, based on the files we obtained, there appear to be only \textbf{7,185 unique} books (excluding \texttt{romance-all.txt} and \texttt{adventure-all.txt} as explained in \ref{sec-original-bc}). We identified potential duplicates based on file names, which suggested that 2,930 books may be duplicated. Using the \texttt{diff} Unix program, we confirmed that BookCorpus contained duplicate, identical text files for all but five of these books. We manually inspected the five exceptions:
\begin{itemize}
\item \texttt{299560.txt} (Third Eye Patch), for which slightly different versions appeared in the ``Thriller'' and ``Science Fiction'' genre folders (only 30 lines differed)
\item \texttt{529220.txt} (On the Rocks), for which slightly different versions appeared in the ``Literature'' and ``Science Fiction'' genre folders (only the title format differed)
\item \texttt{Hopeless-1.txt}, for which identical versions appeared in the ``New Adult'' and ``Young Adult'' genre folders, and a truncated version appeared in the ``Romance'' folder (containing 30\% of the full word count)
\item \texttt{u4622.txt}, for which identical versions appeared in the ``Romance'' and ``Young Adult'' genre folders, and a slightly different version appeared in the ``Science Fiction'' folder (only 15 added lines)
\item \texttt{u4899.txt}, for which a full version appeared in the ``Young Adult'' folder and a truncated version (containing the first 28 words) appeared in the ``Science Fiction'' folder
\end{itemize}
Combined with the \texttt{diff} results, our manual inspection confirmed that each filename represents one unique book, thus BookCorpus contained at most 7,185 unique books.

\subsubsection{Does BookCorpus contain all possible instances (books) or is it a sample?}
\textbf{Sample}. BookCorpus contains free books from smashwords.com which are at least 20,000 words long. Based on metrics from Smashwords \cite{coker2014smashwords}, 11,038 books (as reported in the original BookCorpus dataset) would have represented approximately 3\% of the 336,400 books published on Smashwords as of 2014, while the 7,185 unique books we report would have represented \textbf{2\%}. For reference, as of 2013, the Library of Congress contained 23,592,066 cataloged books \cite{fischer2014library}. We return to the implications of this sample in the discussion (section \ref{sec-discussion}).

\subsubsection{What data does each instance (book) consist of?}
Each book in BookCorpus simply includes the full text from the ebook (often including preamble, copyright text, etc.). However, in research that uses BookCorpus, authors have applied a range of different encoding schemes that change the definition of an ``instance'' (e.g. in GPT-N training, text is encoded using byte-pair encoding).

\subsubsection{Is there a label or target associated with each instance (book)?}
\textbf{No}. The text from each book was originally used for unsupervised training by Zhu and Kiros et al. \cite{zhu2015aligning}, and the only label-like attribute is the genre associated with each book, which is provided by Smashwords.

\subsubsection{Is any information missing from individual instances (books)?}
\textbf{Yes}. We found 98 empty book files in the folder downloaded from the paper's website \cite{zhu2015aligning}. Also, while the authors collected books longer than 20,000 words, we found that 655 files were shorter than 20,000 words, and 291 were shorter than 10,000 words, suggesting that many book files were significantly truncated from their original text.

\subsubsection{Are relationships between individual instances (books) made explicit?}
\textbf{No}. Grouped into folders by genre, the data implicitly links books in the same genre. We also found that duplicate books are implicitly linked through identical filenames. However, no other relationships are made explicit, such as books by the same author, books in the same series, books set in the same context, books addressing the same event, and/or books using the same characters.

\subsubsection{Are there recommended data splits?}
\textbf{No}. The authors use all books in the dataset for unsupervised training, with no splits or subsamples.

\subsubsection{Are there any errors, sources of noise, or redundancies in BookCorpus?} \textbf{Yes}. While some book files appear to be cleaned of preamble and postscript text, many files still contain this text and various other sources of noise. Of particular concern is that we found many copyright-related sentences, for example:
\begin{itemize}
\item ``if you're reading this book and did not purchase it, or it was not purchased for your use only, then please return to smashwords.com and purchase your own copy.'' (n=788)
\item ``this book remains the copyrighted property of the author, and may not be redistributed to others for commercial or non-commercial purposes...'' (n=111)
\item ``although this is a free book, it remains the copyrighted property of the author, and may not be reproduced, copied and distributed for commercial or non-commercial purposes.'' (n=109)
\item ``thank you for respecting the author's work'' (n=70)
\item ``no part of this publication may be copied, reproduced in any format, by any means, electronic or otherwise, without prior consent from the copyright owner and publisher of this book'' (n=16)
\end{itemize}

Here, we note that these sentences represent noise and redundancy, though we return to the issue of copyrights in section \ref{sec-copyright}. As previously noted, BookCorpus also contains many duplicate books: of the 7,185 unique books in the dataset, 2,930 occurred more than once. Most of these (N=2,101) books appeared twice, though many were duplicated multiple times, including some books (N=6) with five copies in BookCorpus. See Table \ref{table-duplicates}.

\begin{table}[]
\begin{small}
\begin{tabular}{|l|l|}
\hline
\textbf{Copies} & \textbf{Number of Books} \\ \hline
1               & 4,255                     \\ \hline
2               & 2,101                     \\ \hline
3               & 741                      \\ \hline
4               & 82                       \\ \hline
5               & 6                        \\ \hline
\end{tabular}
\end{small}
\caption{Number of unique books with different numbers of copies in BookCorpus. 4,255 books only had one copy in BookCorpus (i.e. not duplicated), 2,101 had two copies, etc.}
\label{table-duplicates}
\end{table}

\subsubsection{Is BookCorpus self-contained?}
\textbf{No}. Although Zhu and Kiros et al. \cite{zhu2015aligning} maintained a self-contained version of BookCorpus on their website for some time, there is no longer an ``official,'' publicly-available version. While we were able to obtain the dataset from their website through a security vulnerability, the public web page about the project now states: ``Please visit smashwords.com to collect your own version of BookCorpus'' \cite{zhu2015aligning}. Thus, researchers who wish to use BookCorpus or a similar dataset must either use a new public version such as BookCorpusOpen \cite{bookcorpusopencard}, or generate a new dataset from Smashwords via ``Homemade BookCorpus'' \cite{soskkobayashi2018bookcorpus}.

Smashwords is an ebook website that describes itself as ``the world's largest distributor of indie ebooks.''\footnote{\url{https://www.smashwords.com/}} Launched in 2008 with 140 books and 90 authors, by 2014 (the year before BookCorpus was published) the site hosted 336,400 books from 101,300 authors \cite{coker2014smashwords}. In 2020, it hosted 556,800 books from 154,100 authors \cite{coker2020smashwords}.

\subsubsection{Does BookCorpus contain data that might be considered confidential?} \textbf{Likely no.} While we did find personal contact information in the data (see \ref{subsec-pii}), the books do not appear to contain any other restricted information, especially since authors opt-in to publishing their books.

\subsubsection{Does BookCorpus contain data that, if viewed directly, might be offensive, insulting, threatening, or might otherwise cause anxiety?}
\textbf{Yes}. While this topic warrants further research, as preliminary supporting evidence, we found that 537,878 unique sentences (representing 984,028 total occurrences) in BookCorpus contained one or more words in a commonly-used list of ``Dirty, Naughty, Obscene, and Otherwise Bad Words'' \cite{emerick2018list}. Inspecting a random sample of these sentences, we found they include some fairly innocuous profanities (e.g. the sentence ``oh, shit.'' occurred 250 times), some pornographic dialogue, some hateful slurs, and a range of other potentially problematic content. Again, further research is necessary to explore these sentences, especially given that merely using one of these words does not constitute an offensive or insulting sentence. In section \ref{sec-discussion} we further discuss how some sentences and books may be problematic for various use cases.

\subsubsection{Does BookCorpus relate to people?}
\textbf{Yes}, each book is associated with an author.

\subsubsection{Does BookCorpus identify any subpopulations?}
\textbf{No}. BookCorpus does not identify books by author or any author demographics, and the \texttt{books\_in\_sentences} folder even aggregates all books into just two files. The \texttt{books\_txt\_full} folder identifies 16 genres, though we do not consider genres to be subpopulations since they group books rather than authors.

\subsubsection{Is it possible to identify individuals (i.e., one or more natural persons), either directly or indirectly (i.e., in combination with other data) from BookCorpus?}\label{subsec-pii}
\textbf{Likely yes}. In reviewing a sample of books, we found that many authors provide personally-identifiable information, often in the form of a personal email address for readers interested in contacting them.

\subsubsection{Does the dataset contain data that might be considered sensitive in any way?}
\textbf{Yes}. The aforementioned contact information (email addresses) is sensitive personal information.

\subsubsection{[+] How does the sample compare to the population in terms of genres?}
Compared to BookCorpusOpen and all books on smashwords.com, BookCorpus appears to exhibit several sampling skews. This is to be expected given the filtering applied (only free books, longer than 20,000 words), although some aspects of the skew call for further research attention. See Table \ref{table-genres}.

\begin{table}[]
\begin{small}
\begin{tabular}{|l|l|l|l|}
\hline
\textbf{}        & \textbf{BookCorpus} & \textbf{BookCorpusOpen} & \textbf{Smashwords21} \\ \hline
Romance          & 26.1\% (2880)       & 18.0\% (3314)           & 16.0\% (66083)        \\ \hline
Fantasy          & 13.6\% (1502)       & 17.2\% (3171)           & 10.6\% (44032)        \\ \hline
Science Fiction & 7.5\% (823)         & 13.3\% (2453)           & 7.8\% (32063)         \\ \hline
New Adult       & 6.9\% (766)         & 0.9\% (175)             & 0.7\% (2902)          \\ \hline
Young Adult     & 6.8\% (748)         & 9.5\% (1748)            & 4.6\% (19015)         \\ \hline
Thriller         & 5.9\% (646)         & 7.4\% (1368)            & 5.7\% (23587)         \\ \hline
Mystery          & 5.6\% (621)         & 5.3\% (987)             & 4.7\% (19351)         \\ \hline
Vampires         & 5.4\% (600)         & 0.0\% (0)               & 0.0\% (0)             \\ \hline
Horror           & 4.1\% (448)         & 3.9\% (727)             & 3.9\% (15944)         \\ \hline
Teen             & 3.9\% (430)         & 9.5\% (1752)            & 4.6\% (19154)         \\ \hline
Adventure        & 3.5\% (390)         & 11.5\% (2117)           & 7.1\% (29474)         \\ \hline
Other            & 3.3\% (360)         & 0.1\% (18)              & 0.3\% (1075)          \\ \hline
Literature       & 3.0\% (330)         & 3.0\% (560)             & 2.6\% (10592)         \\ \hline
Humor            & 2.4\% (265)         & 4.1\% (749)             & 3.0\% (12333)         \\ \hline
Historical       & 1.6\% (178)         & 4.7\% (864)             & 4.5\% (18815)         \\ \hline
Themes           & 0.5\% (51)          & 1.3\% (243)             & 1.5\% (6179)          \\ \hline
\end{tabular}
\end{small}
\caption{The distribution of genres represented in the BookCorpus sample, compared to books in the new BookCorpusOpen dataset and all books listed on smashwords.com as of April 2021. Smashwords21 does not contain duplicates (based on book URLs), though BookCorpus and BookCorpusOpen do contain duplicates.}
\label{table-genres}
\end{table}

\subsubsection{[+] How does the sample compare to the population in terms of religious viewpoint?} Further work is needed to thoroughly address this question, as we have not yet generated the appropriate metadata for books in BookCorpus. Furthermore, the books for which we do have metadata (BookCorpusOpen and Smashwords21) only include religions as subjects, not necessarily as viewpoints. For example, metadata might indicate a book is about Islam, though its author writes from an Atheist viewpoint. Despite these limitations, we did find notable skews in religious representation in Smashwords21 and BookCorpusOpen. Following the recently-introduced BOLD framework \cite{dhamala2021bold}, we tabulated based on seven of the most common religions in the world: Sikhism, Judaism, Islam, Hinduism, Christianity, Buddhism, and Atheism. Overall, smashwords.com appears to over-represent books about Christianity, though BookCorpusOpen over-represents books about Islam. See Table \ref{table-religions}.

\begin{table}[]
\begin{small}
\begin{tabular}{|l|l|l|}
\hline
\textbf{}    & \textbf{BookCorpusOpen} & \textbf{Smashwords21} \\ \hline
Sikhism      & 0                       & 15                    \\ \hline
Judaism      & 18                      & 371                   \\ \hline
Islam        & 229                     & 1305                  \\ \hline
Hinduism     & 12                      & 261                   \\ \hline
Christianity & 154                     & 2671                  \\ \hline
Buddhism     & 32                      & 512                   \\ \hline
Atheism      & 18                      & 175                   \\ \hline
\end{tabular}
\end{small}
\caption{Religious subject tally, for books with religious metadata in BookCorpusOpen (N=18,451) and Smashwords21 (N=411,826). Overall, smashwords.com over-represents books about Christianity, though books about Islam are over-represented in the BookCorpusOpen sample.}
\label{table-religions}
\end{table}


\subsection{Collection Process}
\subsubsection{How was the data associated with each instance (book) acquired?} The text for each book was downloaded from smashwords.com.

\subsubsection{What mechanisms or procedures were used to collect BookCorpus?}
The data was collected via scraping software. While the original scraping program is not available, replicas (e.g. \cite{soskkobayashi2018bookcorpus}) operate by first scraping smashwords.com to generate a list of links to free ebooks, downloading each ebook as an epub file, then converting each epub file into a plain text file.

\subsubsection{What was the sampling strategy for BookCorpus?} Books were included in the original BookCorpus if they were available for free on smashwords.com and longer than 20,000 words, thus representing a non-probabilistic \textbf{convenience sample}. The 20,000 word cutoff likely comes from the Smashwords interface, which provides a filtering tool to only display books ``Over 20K words.''

\subsubsection{Who was involved in collecting BookCorpus and how were they compensated?} \textbf{Unknown}. The original paper by Zhu and Kiros et al. \cite{zhu2015aligning} does not specify which authors collected and processed the data, nor how they were compensated.

\subsubsection{Over what timeframe was BookCorpus collected?} \textbf{Unknown}. BookCorpus was originally collected some time before the original paper \cite{zhu2015aligning} was presented at the International Conference on Computer Vision (ICCV) in December 2015.\footnote{\url{http://pamitc.org/iccv15/}} 

\subsubsection{Were any ethical review processes conducted?} \textbf{Likely no}. Zhu and Kiros et al. \cite{zhu2015aligning} do not mention an Institutional Review Board (IRB) or other ethical review process involved in their original paper.

\subsubsection{Does the dataset relate to people?} \textbf{Yes}, each book is associated with an author (thus determining that the following three questions should be addressed).

\subsubsection{Was BookCorpus collected from individuals, or obtained via a third party?} \textbf{Third party}. BookCorpus was collected from smashwords.com, not directly from the authors.

\subsubsection{Were the authors notified about the data collection?} \textbf{Likely no}. Discussing BookCorpus in 2016, Richard Lea wrote in \textit{The Guardian} that ``The only problem is that [researchers] didn’t ask'' \cite{lea2016google}. When notified about BookCorpus and its uses, one author from Smashwords said ``it didn’t even occur to me that a machine could read my book'' \cite{lea2016google}.

\subsubsection{Did the authors consent to the collection and use of their books?} \textbf{No}. While authors on smashwords.com published their books for free, they did not consent to including their work in BookCorpus, and many books contain copyright restrictions intended to prevent redistribution. As described by Richard Lea in \textit{The Guardian} \cite{lea2016google}, many books in BookCorpus include:
\begin{quote}
    a copyright declaration that reserves “all rights”, specifies that the ebook is “licensed for your personal enjoyment only”, and offers the reader thanks for “respecting the hard work of this author”
\end{quote}

Considering these copyright declarations, authors did not explicitly consent to include their work in BookCorpus or related datasets. Using the framework of consentful tech \cite{lee2017consent}, a consentful version of BookCorpus would ideally involve author consent that is \textbf{F}reely given, \textbf{R}eversible, \textbf{I}nformed, \textbf{E}nthusiastic, and \textbf{S}pecific (FRIES).

\subsubsection{Were the authors provided with a mechanism to revoke their consent in the future or for certain uses?} \textbf{Likely no}. For example, if an author released a book for free before BookCorpus was collected, then changed the price and/or copyright after BookCorpus was collected, the book likely remained in BookCorpus. In fact, preliminary analysis suggests that this is the case for at least 438 books in BookCorpus which are no longer free to download from Smashwords, and would cost \$1,182.21 to purchase as of April 2021.

\subsubsection{Has an analysis of the potential impact of BookCorpus and its use on data subjects been conducted?} \textbf{Likely no}. Richard Lea interviewed a handful of authors represented in BookCorpus \cite{lea2016google}, but we are not aware of any holistic impact analysis.

\subsection{Cleaning and Labeling}

\subsubsection{Was any labeling done for BookCorpus?} While the original paper by Zhu and Kiros et al. \cite{zhu2015aligning} did not use labels for supervised learning, each book is labeled with genres. It appears genres are supplied by authors themselves.


\subsubsection{Was any cleaning done for BookCorpus?} \textbf{Likely yes}. The \texttt{.txt} files in BookCorpus seem to have been partially cleaned of some preamble text and postscript text, however, Zhu and Kiros et al. \cite{zhu2015aligning} do not mention the specific cleaning steps. Also, many files still contain some preamble and postscript text, including many sentences about licensing and copyrights. For example, the sentence ``please do not participate in or encourage piracy of copyrighted materials in violation of the author's rights'' occurs at least 40 times in the BookCorpus \texttt{books\_in\_sentences} files.

Additionally, based on samples we reviewed from the original BookCorpus, the text appears to have been tokenized to some degree (e.g. contractions are split into two words), though we were unable to identify the exact procedure.

\subsubsection{Was the ``raw'' data saved in addition to the cleaned data?} \textbf{Unknown}.

\subsubsection{Is the software used to clean BookCorpus available?} While the original software is not available, replication attempts provide some software for turning \texttt{.epub} files into \texttt{.txt} files and subsequently cleaning them.

\subsection{Uses}
\subsubsection{For what tasks has BookCorpus been used?}
BookCorpus was originally used to train sentence embeddings for a system meant to provide descriptions of visual content (i.e. to ``align'' books and movies), but the dataset has since been applied in many different use cases. Namely, BookCorpus has been used to help train more than thirty influential language models \cite{bookcorpuscard}, including Google's enormously influential BERT model which was shown to be applicable to a wide range of language tasks (e.g. answering questions, language inference, translation, and more). 

\subsubsection{Is there a repository that links to any or all papers or systems that use BookCorpus?} On the dataset card for BookCorpus \cite{bookcorpuscard}, Hugging Face provides a list of more than 30 popular language models that were trained or fine-tuned on the dataset.

\subsubsection{What (other) tasks could the dataset be used for?}
Given that embedding text and training language models are useful prerequisites for a huge number of language related tasks, the BookCorpus dataset could in theory be used as part of the pipeline for almost any English language task. However, as discussed below, this work highlights the need for caution when applying this dataset.

\subsubsection{Is there anything about the composition of BookCorpus or the way it was collected and preprocessed/cleaned/labeled that might impact future uses?}
\textbf{Yes}. At the very least, the duplicate books and sampling skews should guide any future uses to curate a subsample of BookCorpus to better serve the task at hand.

\subsubsection{Are there tasks for which BookCorpus should not be used?}
We leave this question to be more thoroughly addressed in future work. However, our work strongly suggests that researchers should use BookCorpus with caution for any task, namely due to potential copyright violations, duplicate books, and sampling skews.

\subsection{Distribution}

\subsubsection{How was BookCorpus originally distributed?} For some time, Zhu and Kiros et al. \cite{zhu2015aligning} distributed BookCorpus from a web page. The page now states ``Please visit smashwords.com to collect your own version of BookCorpus'' \cite{zhu2015aligning}.

\subsubsection{How is BookCorpus currently distributed?} While there have been various efforts to replicate BookCorpus, one of the more formal efforts is BookCorpusOpen \cite{bookcorpusopencard}, included in the Pile \cite{gao2020pile} as ``BookCorpus2.'' Furthermore, GitHub users maintain a ``Homemade BookCorpus'' repository \cite{soskkobayashi2018bookcorpus} with various pre-compiled tarballs that contain thousands of pre-collected books.

\subsubsection{Is BookCorpus distributed under a copyright or other intellectual property (IP) license, and/or under applicable terms of use (ToU)?}\label{sec-copyright} To our knowledge, BookCorpus dataset has never stated any copyright restrictions, however, the same is not true of books within BookCorpus.

In reviewing sources of noise in BookCorpus, we found 111 instances of the sentence, ``this book remains the copyrighted property of the author, and may not be redistributed to others for commercial or non-commercial purposes.'' We also found 109 instances of the sentence ``although this is a free book, it remains the copyrighted property of the author, and may not be reproduced, copied and distributed for commercial or non-commercial purposes.'' This initial analysis makes clear that the distribution of BookCorpus violated copyright restrictions for many books, though further work from copyright experts will be important for clarifying the nature of these violations.

Also, some books in BookCorpus now cost money even though they were free when the original dataset was collected. By matching metadata from Smashwords for 2,680 of the 7,185 unique books in BookCorpus, we found that 406 of these 2,680 books now cost money to download. The total cost to purchase these books as of April 2021 would be \$1,182.21, and this represents a lower bound since we only matched metadata for 2,680 of the 7,185 books in BookCorpus.

\subsubsection{Have any third parties imposed restrictions on BookCorpus?} \textbf{Likely no}.

\subsubsection{Do any export controls or other regulatory restrictions apply to the dataset or to individual instances?} \textbf{Likely no}, notwithstanding the aforementioned copyright restrictions.

\subsection{Maintenance and Evolution}

\subsubsection{Who is supporting/hosting/maintaining BookCorpus?} BookCorpus is not formally maintained or hosted, although a new version called BookCorpusOpen \cite{bookcorpusopencard} was collected by Shawn Presser and included in the Pile \cite{gao2020pile}. As BookCorpus is no longer officially maintained, we answer the below questions by focusing on how other researchers have replicated and extended the BookCorpus data collection approach.

\subsubsection{Is there an erratum for BookCorpus?} \textbf{No}. 
To our knowledge, Zhu and Kiros et al. \cite{zhu2015aligning} have not published any list of corrections or errors in BookCorpus.

\subsubsection{Will BookCorpus be updated?} An updated version of BookCorpus is available as BookCorpusOpen \cite{bookcorpusopencard}. This updated version was published by Presser, not Zhu and Kiros et al. \cite{zhu2015aligning} who created the original BookCorpus.

\subsubsection{Will older versions of BookCorpus continue to be supported/hosted/maintained?} BookCorpus is no longer available from the authors' website, which now tells readers to ``visit smashwords.com to collect your own version of BookCorpus'' \cite{zhu2015aligning}.

\subsubsection{If others want to extend/augment/build on/contribute to BookCorpus, is there a mechanism for them to do so?} \textbf{Yes}, GitHub users maintain a ``Homemade BookCorpus'' repository \cite{soskkobayashi2018bookcorpus} that includes software for collecting books from smashwords.com

\subsubsection{How has BookCorpus been extended/augmented?} The most notable extension of BookCorpus is BookCorpusOpen \cite{bookcorpusopencard}, which was included in ``The Pile'' \cite{gao2020pile} as BookCorpus2, and includes free books from Smashwords as of August 2020.

\section{Discussion}\label{sec-discussion}
This work provides a retrospective datasheet for BookCorpus, as a means of addressing documentation debt for one widely-used machine learning dataset. The datasheet identifies several areas of immediate concern (e.g. copyright violations, duplicated books, and genre skews), as well as other potentially concerning areas that call for future work (e.g. problematic content, skewed religious viewpoints, and lopsided author contributions). Broadly, we suggest that BookCorpus serves as a useful case study for the machine learning and data science communities with regard to documentation debt and dataset curation. Before discussing these broader implications, it is important to note some limitations of our work.

\subsection{Limitations}
While this work addresses all suggested questions for a datasheet \cite{gebru2018datasheets}, it suffers some notable limitations. First, while we obtained BookCorpus data files directly from the original authors’ website \cite{zhu2015aligning}, it remains ambiguous whether these files represent a specific version of BookCorpus, when that version came to exist, and whether it was the version that other researchers used to train models like BERT. For example, some of the empty files in the dataset may reflect books that the researchers removed at some point after the initial data collection. On the other hand, the files we obtain aligned perfectly with many metrics reported by the authors when introducing the dataset, so it is likely that we analyzed either the truly original version of BookCorpus or a very lightly-modified version.

A second limitation is that much of our work represents surface-level analysis, and does not completely ``pay off'' the documentation debt incurred for BookCorpus. Surface-level analysis is often sufficient to reveal important areas of concern (as this work demonstrates), however, it also means that some areas need more thorough analysis and attention. We now identify some areas in both of these categories, which we plan to explore further in future work.

\subsection{Areas of Immediate Concern}
This work identified at least three immediate areas of concern with respect to BookCorpus: copyright violations, duplicate books, and genre skews. In terms of copyright violations, we found that many books contained copyright claims that should prevent distribution in the form of a free machine learning dataset. Many books explicitly claimed that they ``may not be redistributed to others for commercial or non-commercial purposes,'' and thus should not have been included in BookCorpus. Also, at least 406 books were included in BookCorpus for free even though the authors have since increased the price of the book. For example, the full text of \textit{Prodigy} by Edward Mullen is in BookCorpus (as \texttt{366549.txt}), even though the author now charges \$1.99 to download the book from Smashwords \cite{mullen2013prodigy}.

A second immediate area of concern is the duplication of books. BookCorpus is often cited as containing 11,038 books, though this work finds that only 7,185 of the books are unique. The duplicate books did not necessarily impede BookCorpus' original use case \cite{zhu2015aligning}, however, redundant text has been a key concern in improving training datasets for language models. The Colossal Clean Crawled Corpus (C4) \cite{raffel2019exploring}, for example, discards all but one of any three-sentence span that occurred more than once. Future research using BookCorpus should take care to address duplicate books and sentences.

A final area of concern is the skewed genre representation we identified in BookCorpus, which over-represented romance books. This skew may emerge from a broader pattern in the self-publishing ebook industry, where authors consistently find that ``kinky'' romance novels are especially lucrative \cite{jeong2018romance,flood2019plagiarism,samuels2018amazon}. In other words, because romance is a dominant genre in the set of self-published ebooks, romance is also dominant in the subset of free ebooks.

But romance novels often contain explicit content that can be problematic with respect to many use cases for BookCorpus, particularly when context is ignored. For example, BookCorpus contains a book called ``The Cop and the Girl from the Coffee Shop'' (\texttt{308710.txt}) \cite{towers2013cop}, which notes in the preamble that ``The material in this book is intended for ages 18+ it may contain adult subject matter including explicit sexual content, profanity, drug use and violence.'' On smashwords.com, the book is tagged with ``alpha male,'' and ``submissive female,'' and thus could contribute to well-documented gender discrimination in computational language technologies \cite{bolukbasi2016man,webster2018mind,caliskan2017semantics}. Little harm is done when mature audiences knowingly consume adult content, but this awareness is often not the case for text generated by language models. Thus, while the genre skew is concerning in and of itself, this example of adult content also highlights a concerning area that calls for future work.

\subsection{Areas for Future Work}
Overall there are many potential directions for research in dataset documentation, though here we note three that surfaced in our work: problematic content, skews in religious viewpoints, and lopsided author contributions. The book mentioned above, ``The Cop and the Girl from the Coffee Shop,'' represents just one example of content that would likely impede language technology in many use cases. For example, a generative language model trained on many similar books would be susceptible to generating pornographic text and reproducing harmful gender stereotypes. That is to say: even though the original text may have been written in good faith and consumed by informed, consenting adults, feeding this text to a language model could easily produce similar text in very different contexts. However, this is just one book in the dataset, and further work is needed to determine the extent of this potentially problematic content within BookCorpus.

Further work is also needed to clarify skews in religious viewpoint. Metrics from BOLD \cite{dhamala2021bold} suggest that some language models trained on BookCorpus favor Christianity (based on sentiment analysis), and our Smashwords21 dataset does suggest that Smashwords over-represents books about Christianity. However, we do not yet have the metadata needed to precisely determine religious representation in BookCorpus. Further work is also needed to potentially distinguish between books \textit{about} a given religion and books written from a particular religious viewpoint.

Finally, future work should delve further into measuring lopsided author contributions. Once again our Smashwords21 dataset points to several points of potential concern, such as ``super-authors'' that publish hundreds of books. This prompts normative considerations about what an ideal ``book'' dataset \textit{should} look like: which writers should these datasets contain? Should work by prolific writers be sub-sampled? If so, how?

We suggest that machine learning research will greatly benefit from engaging these questions, pursuing more  detailed dataset documentation, and developing tools to inspect datasets.

\subsection{Conclusion}
This work begins to pay off some of the ``documentation debt'' for machine learning datasets. We specifically address BookCorpus, highlighting a number of immediate concerns and important areas for future work. Our findings suggest that BookCorpus provides a useful case study for the machine learning and data science communities, showing that widely-used datasets can have worrisome attributes when sparsely documented. Some may suggest that sophisticated language models, strategic fine-tuning, and/or larger datasets can drown out any effects of the worrisome attributes we highlight in this work. However, datasets themselves remain the most ``upstream'' factor for improving language models, embedding tools, and other language technologies, and researchers should act accordingly.

The NeurIPS ``Datasets and Benchmarks'' track \cite{Vanschoren2021announcing} shows that the community has started to recognize the importance of well-documented datasets. NeurIPS has also published guidelines suggesting that authors provide dataset documentation when submitting papers \cite{Beygelzimer2021introducing}, ideally reducing the need for retrospective documentation efforts. In the meantime, post hoc efforts (like the one offered in this paper) provide a key method for understanding and improving the datasets that power machine learning.

\begin{acks}
Thanks to the Computational Journalism Lab for helpful comments and questions during a lab presentation.
\end{acks}

\bibliographystyle{ACM-Reference-Format}
\bibliography{sources}





\end{document}